\documentclass[journal]{IEEEtran}
\usepackage{ifpdf}
\usepackage{cite}
\usepackage[pdftex]{graphicx}
\usepackage[cmex10]{amsmath}
\usepackage{amsfonts}
\usepackage{algorithmic}
\usepackage{array}
\usepackage{fixltx2e}
\usepackage{stfloats}
\usepackage{url}
\usepackage{multirow}

\begin{document}
\title{Mask-aware networks for crowd counting}

\author{Shengqin Jiang,
        Xiaobo Lu,
        Yinjie Lei,
        Lingqiao Liu

\thanks{This work is done when the first author visits the University of Adelaide.}% <-this % stops a space

\thanks{S. Jiang is with the School of Automatic, Southeast University, Nanjing 210096, China; the School of Computer Science, University of Adelaide, Adelaide, SA 5005, Australia; the Key Laboratory of Measurement and Control of Complex Systems of Engineering, Ministry of Education, Nanjing 210096, China e-mail: (jiangshengmeng@126.com).}% <-this % stops a space
\thanks{X. Lu are with the School of Automatic, Southeast University, Nanjing 210096, China; Key Laboratory of Measurement and Control of Complex Systems of Engineering, Ministry of Education, Nanjing 210096, China e-mail: (xblu2013@126.com).}% <-this % stops a space
\thanks{Y. Lei is with the College of Electronics and Information Engineering, Sichuan University, Chengdu 610064, China (e-mail: yinjie@scu.edu.cn).} % <-this % stops a space
\thanks{L. Liu is with the School of Computer Science, University of Adelaide,
Adelaide, SA 5005, Australia (e-mail: lingqiao.liu@adelaide.edu.au).}% <-this % stops a space
}

\markboth{IEEE Transactions on Circuits and Systems for Video Technology}%
{Jiang \MakeLowercase{\textit{et al.}}: Mask-aware networks for crowd counting}

\maketitle

\begin{abstract}
Crowd counting problem aims to count the number of objects within an image or a frame in the videos and is usually solved by estimating the density map generated from the object location annotations. The values in the density map, by nature, take two possible states: zero indicating no object around, a non-zero value indicating the existence of objects and the value denoting the local object density. In contrast to traditional methods which do not differentiate the density prediction of these two states, we propose to use a dedicated network branch to predict the object/non-object mask and then combine its prediction with the input image to produce the density map. Our rationale is that the mask prediction could be better modeled as a binary segmentation problem and the difficulty of estimating the density could be reduced if the mask is known. A key to the proposed scheme is the strategy of incorporating the mask prediction into the density map estimator. To this end, we study five possible solutions, and via analysis and experimental validation we identify the most effective one. Through extensive experiments on five public datasets, we demonstrate the superior performance of the proposed approach over the baselines and show that our network could achieve the state-of-the-art performance.
\end{abstract}

\IEEEpeerreviewmaketitle

\section{Introduction}
\IEEEPARstart{C}{rowd} counting is a significant topic for crowd understanding and analysis~\cite{choi2014understanding,shao2017crowded,huang2018body,li2018deep}, which has attracted much attention in multimedia and computer vision community due to large and practical demands for better management, safety and security ~\cite{zhang2016data,sindagi2018survey}. It aims to count the number of objects within an image or a frame in the videos and is a very challenging problem because the objects-of-interest, e.g., people, can occur at a variety of scales, with heavy occlusions and cluttered visual appearances. Also, due to the difficulty in providing highly detailed annotations such as object bounding boxes or instance-level segmentation masks, existing datasets usually adopt a weak-level annotating scheme by labeling each object with a dot inside. These challenges make the traditional detection based approach less robust~\cite{ren2015faster,liu2016ssd,redmon2016you,li2018multistage} and most existing methods~\cite{zhang2016single,wang2015deep,sam2017switching} choose to solve this problem by estimating a density map generated from the dot-level annotation. Once the density map is correctly estimated, the object count can be obtained by simply summing over the density values in the map.

In the current density map annotation scheme, the values in the density map are all non-negative and only pixels close to an annotated dot can have nonzero values. In other words, a density value could exhibit two states: zero indicating no objects within its neighborhood; a non-zero value indicating the existence of objects with the value denoting the local object density. In fact, for the density maps of many images, a significant portion of pixels will only take the zero value.

The above observation suggests that the density map estimation implicitly involves two steps: estimating whether a pixel belongs to the foreground or background (object/non-object) and estimating the density value of the foreground region. Certainly, these two steps can be achieved by a single density map estimator which is trained with the traditional regression objectives, e.g., mean square error, as in the existing approaches. However, in this paper, we argue the benefits of explicitly separating the mask prediction from the density estimation. More specifically, we propose to use a dedicated branch of a network to first predict the foreground/background mask, and then fuse the prediction with the input image to produce the final density map estimation. The motivation of this strategy is that the first step is essentially a binary segmentation problem and it can be better trained with segmentation loss such as cross-entropy loss. On the other hand, conditioned on the prediction of the mask, the estimation of the density map can be simpler than its unconditioned counterpart. Consequently, the overall regression quality could be improved.  

The critical question of the above-proposed process is how to incorporate the mask prediction information into the density map estimator. In this paper, we study five different variants for achieving this incorporation. Specifically, in the proposed five solutions, we consider the following factors and their combinatorial effects: (1) The representation of mask information. Should we use the binary form of the mask prediction or the predicted mask posterior, i.e., the probability of a pixel being the foreground. (2) The way to incorporate the mask information. By simply multiplying the estimated mask or fusing this part of information with a neural network. We analyze, both theoretically and experimentally, the pros and cons of the proposed methods and identify the last one as our best solution.  More specifically, in this solution, we feed the estimated object posterior into a few convolutional layers and together with the information from the input image to produce the final density map. Through extensive experiments on five public datasets, we demonstrate the superior performance of the proposed approach over the competitive baseline and show that our method can achieve the state-of-the-art crowd counting performance. 

In sum, the contributions of this paper are threefold:
\begin{itemize}
\item We propose a strategy to separately model the foreground/background mask with a dedicated neural network branch and training objective.  
\item We study five different solutions of incorporating the mask prediction information into the overall density map estimation and identify the most effective one.
\item The proposed method achieves the state-of-the-art crowd counting performance on various datasets. 
\end{itemize}

\section{Related work}
To date, many approaches have been proposed to study the issues existing in crowd counting~\cite{ge2009marked, ge2012vision, idrees2015detecting, onoro2016towards, shao2017crowded}. Here we present a brief review of the related work. For a more detailed survey of crowd counting, we refer the readers to~\cite{loy2013crowd,tripathi2018convolutional, sindagi2018survey}. %(\cite{Loy2013,Vishwanath, and Vishal 2018).

Detection seems to be a straightforward solution for crowd counting. The most early methods use hand-crafted features such as Haar wavelets~\cite{viola2004robust}, histogram oriented gradients~\cite{dalal2005histograms} to model the pedestrian, which are then fed to classifiers to distinguish whether there is pedestrian or not. Initially,~\cite{dalal2005histograms,enzweiler2008monocular} studied the monocular pedestrian detection by a diverse set of low-level feature-based systems. Although monocular-based methods work well in a low density region, the performance is severely affected when they meet the crowded scenes with occlusion and scene clutter. To further consider this issue, more information of the pedestrian is taken into account. Zhao et al.~\cite{zhao2008segmentation} used multiple partially occluded human hypotheses in a Bayesian framework to build a model-based approach to interpret the image observations. The authors in~\cite{yang2003counting} extracted the foreground and then aggregated the obtained silhouettes over a network to compute bounds about the crowd number and locations. 

Nevertheless, the representation ability of the low-level features is limited, which cannot be applied in many real scenarios. Recently, many approaches resort to application of the CNN-based detectors such as Faster RCNN~\cite{ren2017faster}, YOLO~\cite{redmon2016you}, SSD~\cite{liu2016ssd}, which are trained end to end and have a good generalization compared with the traditional ones. These methods make a great progress in terms of detection performance and speed. However, for a heavily occluded and cluttered scenario, accurately detecting each object instance is still very difficult. 

As a alternative solution of the detection-based methods, the regression-based approaches are proposed to tackle the extremely dense crowds. Initially, these approaches learn a mapping or relation between the features of local patches and the counts. Actually, they avoid learning some independent detectors. For example, the authors~\cite{lempitsky2010learning} proposed to cast the crowd counting problem as a density map estimation problem. The integral of the image density over any image region gives the count of objects within that region. It was shown that the density map regression framework offers a robust crowd counting solution for various challenging scenarios, and since then it becomes the mainstream framework for this problem. Various extensions~\cite{chan2012counting,ma2013crossing,idrees2013multi,chen2013cumulative} have been proposed to further improve the training and prediction of density maps. Ma et al.~\cite{ma2013crossing}  studied an integer programming method for estimating the instantaneous count of pedestrians crossing a line of interest in a video sequence. Idrees et al.~\cite{idrees2013multi} argued that it is not reliable by only using one single feature or detection method for counting task when facing the high-level density crowds. And they also reported that the spatial relationship is an importance information to constrain the counts in neighboring local regions. Chen et al.~\cite{chen2013cumulative} studied the challenges of inconsistent features along with sparse and imbalanced data, and proposed to learn a regression model by using cumulative attribute-based representation.

With the breakthrough of deep learning in the past years, most recent works on crowd counting are based on convolutional neural networks.  the authors in~\cite{wang2015deep} built an end-to-end CNN regression model to count the people in extremely crowd scenes. In the same year,~\cite{zhang2015cross} proposed a deep CNN method, which is trained alternatively with two related learning objectives, crowd density estimation, and crowd count estimation. Later,~\cite{wang2015deep} introduced a CNN architecture that is fed with a whole image and directly outputs the final count. To address the large variations in people or head size,~\cite{zhang2016single} exploited a multi-column neural network (MCNN) by using receptive fields of different sizes in each column. The authors~\cite{sam2017switching} proposed a path switching architecture, called Switching-CNN, to deal with the variation of object density within a scene. In order to gain better performance,~\cite{sindagi2017generating} proposed a Contextual Pyramid CNN by incorporating different levels of contextual information to achieve state-of-the-art performance. At the same time, more recent works ~\cite{huang2018body,ranjan2018iterative,shen2018crowd,shi2018crowd,li2018csrnet} have gained promising results and advanced the development of crowd counting. 

In this paper, we propose a mask-aware network for crowd counting which incorporates the background/foreground mask information into the network for more accurate density regression. In terms of the network architecture design, our network is somehow similar to the recent work ~\cite{sam2018top}, which utilizes the top-down feedback to correct the initial prediction. However, our approach considers the background/foreground mask information and we will show later in the experiments that this consideration is crucial for achieving our good performance. In terms of using mask information, there has been some successful cases in the areas of object segmentation and person re-identification~\cite{he2017mask,lyu2018mask}. However, to our knowledge, our work is the first one that systematically studies the effect of mask-aware networks for crowd counting.

\begin{figure*}[!t]
	\centering
    \includegraphics[scale=0.9]{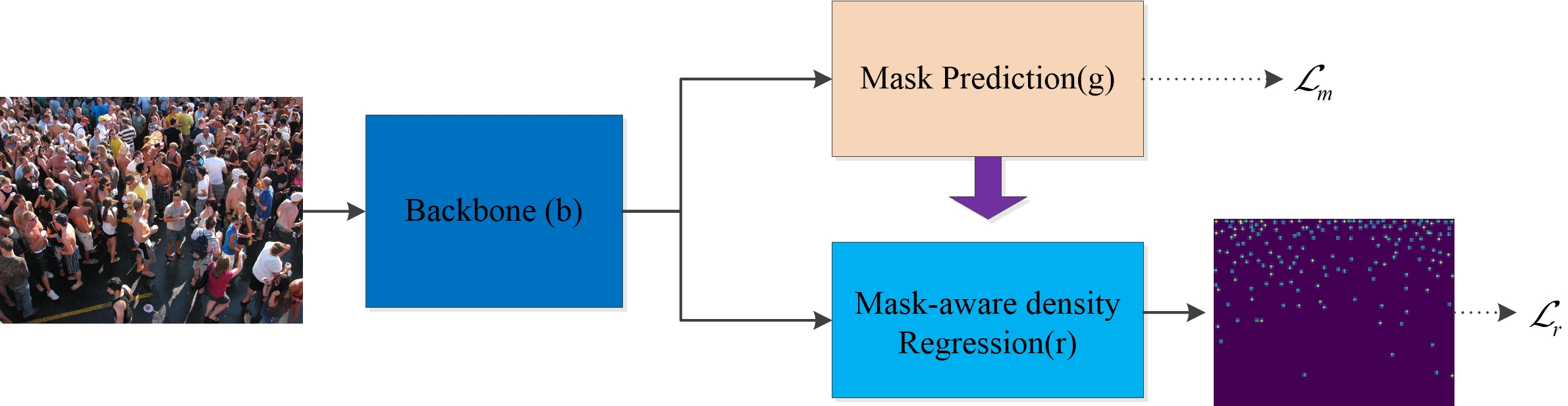}
	\caption{An overview of our proposed method. It contains three modules: the backbone, the mask prediction branch and the mask-aware density regression branch.}
	\label{fig:fig1}
\end{figure*} 

\section{Our proposed method}
\subsection{Density map estimation}
Before elaborating the design of our network, we first briefly introduce the creation of the ground-truth density maps and the related training losses. 

This paper considers the case that a point-wise annotation is provided for training images. Specifically, a dot is annotated within each object-of-interest, i.e., people head. This pointwise annotation is further converted into a density map:
\begin{equation}
d(\mathbf{x}) = \sum_{\mathbf{x}_i\in \mathcal{A}} G(\frac{\|\mathbf{x}-\mathbf{x}_i\|^2}{\sigma^2}),
\end{equation} 

where $\mathbf{x} \in \mathbb{R}^2$denotes the image coordinate and $\mathbf{x}_i$ denotes the annotated head location. $G(\frac{\|\mathbf{x}-\mathbf{x}_i\|^2}{\sigma^2})$ denotes a Gaussian kernel with $\mathbf{x}_i$ as the mean vector and $\sigma^2$ as the empirically chosen variance term. A typical choice of $\sigma^2$ will make $G(\frac{\|\mathbf{x}-\mathbf{x}_i\|^2}{\sigma^2}) = 0$ if $\mathbf{x}$ is not within the local neighborhood of $\mathbf{x}_i$. It is also easy to verify that the integral of $d(\mathbf{x})$ over $\mathbf{x}$ equals to the total number of objects. Thus the counting problem can be cast as a density regression problem and the mean square loss (MSE) is usually employed to train the regressor:

\begin{equation}
	\mathcal{L}_r (\lambda) = \sum_{x} \left(f(\mathbf{x};\lambda) - d(\mathbf{x})\right)^2,
\end{equation}

where $f(\mathbf{x};\lambda)$ is the regression network and $\lambda$ denotes the model parameters. 

\subsection{Method overview}
The overview of the proposed method is shown in Figure~\ref{fig:fig1}. For the clarity of presentation, we divide the network into three parts: a backbone subnetwork $b$, a mask prediction branch $g$ and the mask-aware density regressor $r$. The backbone generates the feature representation of the input image and is shared across all the mask-aware density regressors as discussed below. The mask prediction branch predicts the foreground/background mask. The mask-aware density regressor is the key contribution of this paper and five different designs will be presented in this section. 

As mentioned in the introduction section, the value of $d(\mathbf{x})$ takes two possible states: with a zero value indicating no object around while with a nonzero value indicating the existence of object, and for a large portion of $\mathbf{x}$ its corresponding $d(\mathbf{x})$ is zero. This observation inspires us to design a dedicated branch of a neural network to predict the foreground/background (object/non-object) mask and we train this branch as a binary segmentation network. Then we can utilize the mask prediction information to guide the overall density estimation. Formally this process is denoted as $g((b\circ g)(I), b(I))$, where $I$ denotes the input image and the training objective can be written as:
\begin{equation}
\mathcal{L}_m \left( (b\circ g)(I), M\right) + \alpha \mathcal{L}_r \left( f((b\circ g)(I), b(I)), D \right),
\end{equation}

where $\mathcal{L}_m$ is the loss function for evaluating the performance of mask prediction; $\mathcal{L}_r$ is the loss function for evaluating the overall density estimation; $M, D$ are the groundtruth of the mask and the density map, respectively. The ground-truth mask is defined as $M(\mathbf{x}) = sign(d(\mathbf{x}))$. More specifically,   The mask is used to distinguish the background and foreground, which means the threshold is 0. That is, if the counting number of each pixel is greater than 0, the pixel is then classified to 1 (i.e., foreground), otherwise, 0 (i.e., background); $\alpha$ is a trade-off parameter. 

\subsection{Backbone sub-network}

\begin{figure}
	\centering
    \includegraphics[scale=0.6]{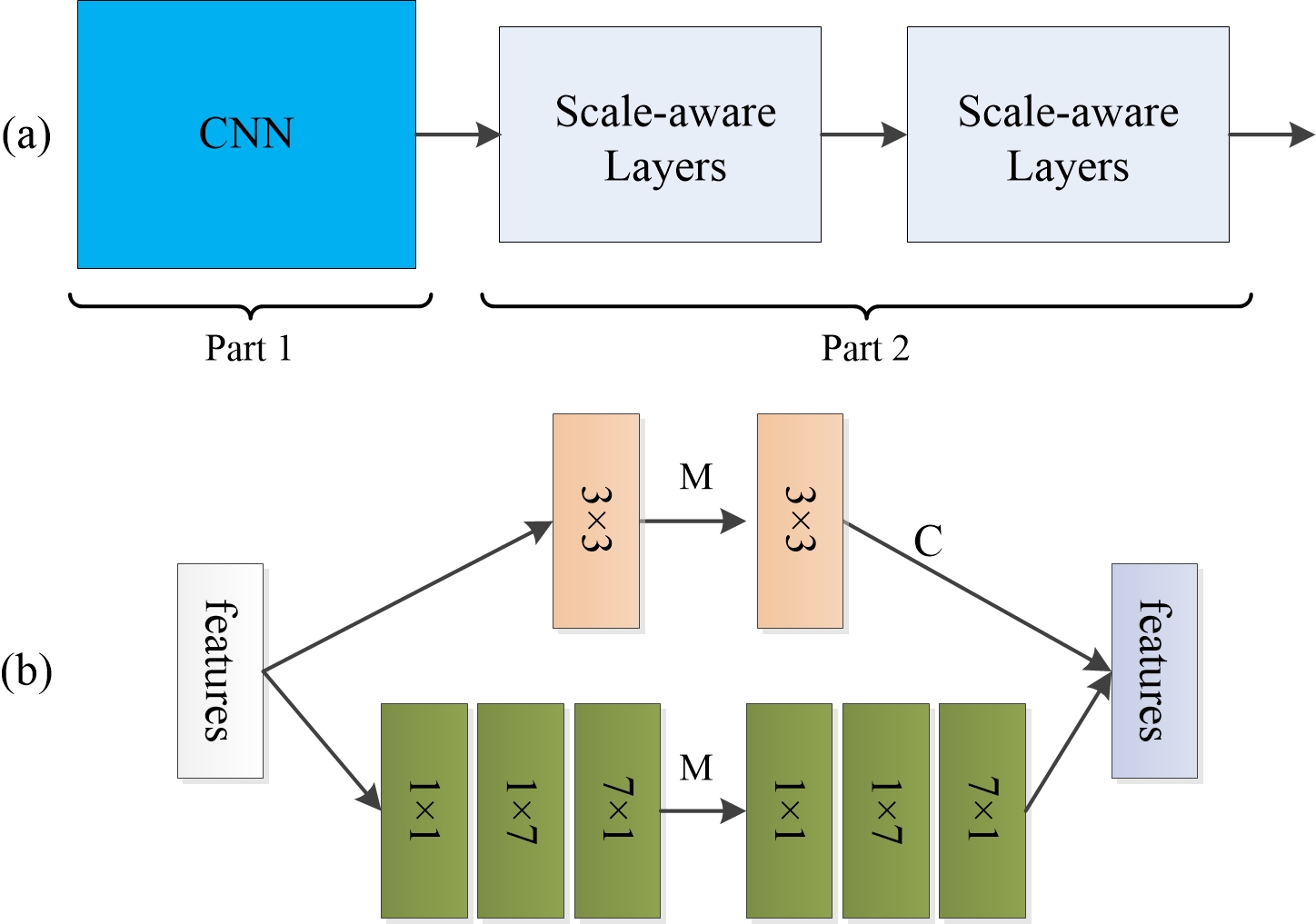}
	\caption{The architecture of the backbone subnetwork. 'M' denotes pooling operation and 'C' denotes concatenation of the features.}
	\label{fig:fig2}
\end{figure}

The architecture of the backbone sub-network is shown in Figure \ref{fig:fig2} (a). It consists of two parts. The first part is a typical multi-layer CNN and the second part is similar to the blocks in the Inception network ~\cite{szegedy2016rethinking}. The layers of first part are C(1, 64, 3)-C(64, 64, 3)-MP-C(64, 128, 3)-C(128, 128, 3)-MP-C(128, 128, 3) where C($x$, $y$, $z$) denotes the convolution layer with $x$ channels of input, $y$ channels of output and $ z\times z$ convolution kernel and MP denotes max pooling. The second part has two identical units (the structure of each unit is shown in Figure \ref{fig:fig2} (b)) and its purpose is to encourage the network using information from different scales. This is in a spirit similar to the design of multi-column CNN (MCNN ~\cite{zhang2016single}). However, our backbone only adopts multiple scale paths at the second part and uses the separable convolution layers($1\times7$ and $7\times1$) as shown in Figure \ref{fig:fig2} (b). One empirically suggests that the backbone is completely superior to MCNN in terms of the performance (as shown in the Part IV). 

The above proposed sub-network is a light-weight strategy which is completely trained from scratch. To further verify the following proposed solution is not specialized for the proposed sub-network, we also employ a pre-trained VGG16 model as our backbone to train our solution followed by the state-of-the-art model CSRNet\cite{li2018csrnet}.

\subsection{Mask prediction branch}
The mask prediction branch consists of multiple convolutional layers. Specifically, the architecture could be denoted as C(256, 256, 3)-C(256, 1, 1). In our implementation, we can train the mask prediction branch with focal loss ~\cite{lin2018focal} as the training objective $ \mathcal{L}_m $. It is calculated by applying the sigmoid function to the output activation of the mask prediction branch. As reported in ~\cite{lin2018focal}, focal loss is designed to tackle the imbalance between foreground and background during training.In most crowd scenarios, there may exist the imbalance issues. But we find it does not make much difference in our experiment when using the focal loss and traditional binary cross-entropy loss, which will be reported in Section IV. Here, we use focal loss as a general setting for cross-entropy loss. That is, $\mathcal{L}_m$ is binary cross-entropy loss when $\gamma=0$.

Note that traditional single branch density map estimation networks still need to determine (although implicitly) whether a pixel belongs to the foreground or background. They achieve this capability by using the MSE loss while our mask prediction branch utilizes the cross-entropy loss (with the focal loss) which is generally considered as a better objective for segmentation tasks.
%However, they are trained based on the MSE loss. The proposed mask prediction branch uses cross-entropy is  

\subsection{Mask-aware density density regressor}

The ways of incorporating the mask prediction information into the density regression are critical in our proposed method. In the following part, we consider five possible solutions. 
\begin{figure}[h]
	\centering
    \includegraphics[scale=0.5]{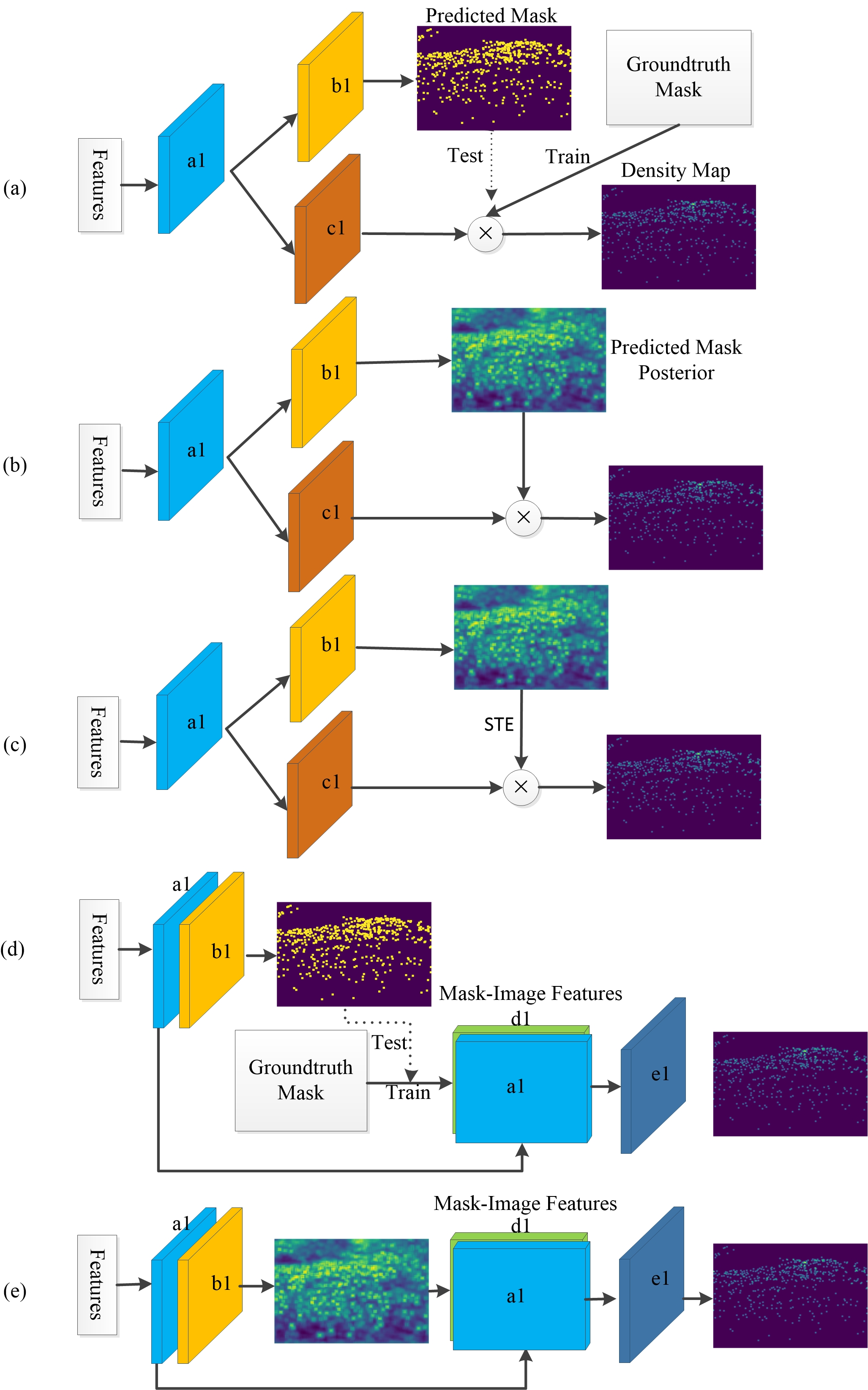}
	\caption{Five different architectures for the mask-aware density regressor. (a), (b) and (c) use element-wise product to incorporate the predicted mask information, where (a) uses the groundtruth mask, (b) directly uses the predicted mask posterior and (c) uses STE function to backpropagate the gradient; (d) and (e) fuse the information from predicted mask by several convolutional layers, where (d) uses groundtruth mask but using predicted mask for test, and (e) learns the mask-image features from the output of mask prediction in an end-to-end fashion.}
	\label{fig:fig3}
\end{figure}

% The 'features' block means the features are obtained by the final Scale-aware Layers shown in Fig. 2; The 'a1' block is a shared convolutional layers(C(512, 256, 3)); The 'b1' block is the density regression layers (C(256, 256, 3)-C(256, 1, 1)); The 'c1' block is the mask prediction layers (C(256, 256, 3)-C(256, 1, 1)); The 'd1' block is the feature maps obtained from the mask prediction (C(1, 512, 3)-C(512, 256, 3)); The 'e1' block is used to regress the final density (C(512, 256, 3)-C(256, 1, 1)); The 'sigmoid' (or 'STE') means the final feature map is processed by function sigmoid (or STE); ' $\times$ ' means dot product.

\noindent\textbf{Solution 1.} By definition, the mask indicates which part of density should be nonzero/zero. Thus a straightforward way to fuse mask information with the density map estimation is to elementwisely multiply the estimated density map by the estimated mask. Our first solution uses this scheme, as shown in Figure \ref{fig:fig3} (a).  At the training stage, the training goal of the mask prediction branch to produce the ground-truth mask. Thus we directly multiply the density map from the density estimation branch with the ground-truth mask at the training stage. Note that this solution essentially requires the density estimation branch only focuses on the estimation of the density in the foreground region. 

While being conceptually straightforward, this solution, however, ignores the possible connection between the mask prediction branch and density estimation branch. Noted that the gradient of the $\mathcal{L}_r$ will not pass through the mask prediction branch at the training time. This suggests that these two branches are essentially trained independently with separated objectives.

%The first solution considered is shown in Figure \ref{fig:fig3} (a). The network contains a density estimation branch which is with the similar structure as the mask prediction branch. The output of the density branch is elementwisely multiplied by the ground-truth mask at the training stage. At the test stage, we do not have access to the ground truth mask, so we will use the predicted mask instead. Note that this solution essentially requires the density estimation branch and only focuses on the estimation of the density in the foreground region. 

%While being conceptually straightforward, this solution, however, ignores the possible connection between the mask prediction branch and density estimation branch. Noted that the gradient of the $\mathcal{L}_r$ will not pass through the mask prediction branch at the training time. This suggests that these two branches are essentially trained independently with separated objectives. 

\noindent \textbf{Solution 2.} To facilitate the connection between the prediction branch and the density estimation branch, we modify solution 1 and propose the second solution as shown in Figure \ref{fig:fig3} (b). The difference is that instead of using the ground-truth mask we use the estimated posterior of the foreground (the soft version of the mask prediction) to multiply the output of the density estimation branch. In this case, the gradient loss $\mathcal{L}_r$ can backpropagate to the mask prediction branch, making it jointly adapt with the density estimation branch to produce the final estimation.

\noindent \textbf{Solution 3.} In solution 2, the final density prediction is the multiplication of the posterior and the output of the density estimation branch. Since the value of the posterior is between 0 and 1. It is not a perfect mask and could make the estimation sensitive to the confidence of mask prediction. To overcome this drawback, we propose to multiply the predicted binary mask instead. The generation of the mask involves a nondifferentiable hard-thresholding operation. To backpropagate the gradient, we adopt the straight-through estimator (STE for short) ~\cite{bengio2013estimating} to approximate this operator as shown in Figure \ref{fig:fig3} (c). Formally, in the forward calculation, the predicted mask used in the multiplication is obtained via $\hat{m_i}= h(p(\mathbf{x}_i))$, where $h(\cdot)$ returns 1 if the $p(\mathbf{x}_i)$ is greater than 0.5, otherwise 0. In backpropagation, we approximate the gradient as 
\begin{equation}
\frac{\partial \hat{m_i}}{\partial(p(\mathbf{x}_i))} \approx 1.
\end{equation} The schematic illustration of this solution is shown in Figure \ref{fig:fig3} (c).

\noindent\textbf{Solution 4.} The above two designs are based on the elementwise product operation to merge the information of mask prediction, which can be quite restrictive and potentially sensitive to the mask prediction quality. Here we propose an alternative solution as shown in Figure \ref{fig:fig3} (d). The idea is to use several convolutional layers to map the mask into a feature map which can be further concatenated with the image features to perform the density estimation. Similar to solution 1, we can use the ground-truth mask at the training time and replace it with the predicted ones at the test stage. In this design, we use one channel of ground truth mask to generate a feature map with 256 channels, and then we concatenate the 256 channels from previous layers as the input for the last density map regressor. Finally, the architecture of the density map regressor is C(512, 256, 3)-C(256, 256, 3)-C(256, 1, 1). 

\noindent\textbf{Solution 5.} Similar to the solution 2, we could further improve solution 4 by using the estimated posterior probability to replace the predicted mask. This allows joint training of all the components of the network. The structure of this solution is shown in Figure \ref{fig:fig3} (e). Since this structure learns the incorporation operation through a set of convolutional layers rather than a simple elementwise product, we postulate that it can be less sensitive to the value of posterior estimation. 

\subsection{Implementation details} 
%The network consists of two parts: backbone and mask-aware density regression. As there are large variations in people or head size, we consider a new scale-aware layers shown in Figure 7. Compared to MCNN, our aware-layers only consider two scales but using shallow 3*3 layers and deep 1*7 and 7*1 layers. This consideration lies in that it reduces the parameter of 7*7 kernel and also provides more abstract information compared with only using 7*7 kernel. We also use two blocks of the scale-aware blocks to percept the scale variations where the second block using Max pooling with padding to keep the resolution of final output.The layers before scale-aware layer are conv3-1-64-conv3-64-64-MP-conv-3-64-128-conv3-128-128-MP-conv3-128-128 where MP is max pooling. The mask-aware density regression has been described in details and so we do not list them here.

Our proposed method is trained from scratch based on the Pytorch framework. Firstly, we generate the Ground truth(GT) following from previous method by using a Gaussian kernel. We fix the kernel size for all datasets to generate the density map although using geometry-adaptive kernel for different datasets might further improve prediction performance.

For the proposed multi-scale backbone shown in Figure \ref{fig:fig2}, we randomly mirror the cropped training images and their associated GT on the fly. What's more, the initialization of the network is drawn from normal distribution with 0.01 standard deviation. In order to gain a quicker training speed, the Adam optimizer is used to train the network before 11\textit{th} epoch and then switch to mini-batch Stochastic Gradient Descent (SGD). The learning rate is initially set to 1e-5 and then is decreased by a factor of 0.1 every 20 epochs.

%For the proposed multi-scale backbone shown in Figure \ref{fig:fig2}, fixed size patches are randomly cropped form each image in the tested dataset to augment the training data. During training, we also randomly mirror the training images and their associated GT on the fly. What's more, the initialization of the network is drawn from normal distribution with 0.01 standard deviation. In order to gain a quicker training speed, the Adam optimizer is used to train the network before 11\textit{th} epoch and then switch to mini-batch Stochastic Gradient Descent (SGD). The learning rate is initially set to 1e-5 and then is decreased by a factor of 0.1 every 20 epochs. %The batch size is set to 1. The learning rate is initially set to 1e-5 and then is decreased by a factor of 0.1 every 20 epochs. % throughout all of the experiments. 

As for using pre-trained VGG16 as backbone, we use original images as training dataset without data augmentation unless otherwise stated. In our experiments, we use SGD optimizer train the network for the datasets with different size of images and the rest ones use Adam optimizer. In addition, we use standard cross-entropy loss for all the experiments.

\section{Experiment}

In this section, we conduct experiments on three challenging public datasets to demonstrate the effectiveness our proposed method including ShanghaiTech dataset ~\cite{zhang2016single}, UCF\_CC\_50 dataset ~\cite{idrees2013multi} and WorldExpo' 10 dataset ~\cite{zhang2015cross}. The purposes of our experiments are threefold: (1) verify if the proposed mask-aware strategies lead to significant improvement over the baselines. (2) identify the most effective mask-aware density estimation solution. (3) compare our proposed approach against the state-of-the-art methods. In our experiments, we use ShanghaiTech dataset A to achieve the first and the second objective. The identified best-performed solution will then be compared against the state-of-the-art on all three datasets. 

In what follows, we present the evaluation criterion and datasets in our experiments. Then we present a detailed analysis of the proposed solutions and identify the most effective one. Finally, we compare our method with other state-of-the-art methods.

\subsection{Evaluation metrics }

We use mean absolute error (MAE) and mean square error (MSE) as the evaluation metrics. The two metrics are defined as follows:

\begin{equation}
\textit{MAE} =  \frac{1}{N}\sum_{i=1}^{N}\left | Pr_i - Gt_i \right |
\end{equation}\\
and
\begin{equation}
\textit{MSE} = \sqrt{\frac{1}{N}\sum_{i=1}^{N}\left ( Pr_i - Gt_i \right )^{^{2}}},
\end{equation} 

where \textit{N} is the number of the images in the test dataset, and $Pr_i$ denotes the predicted object count obtained from the network for the \textit{i-th} image while $Gt_i$ denotes the ground truth count of the \textit{i-th} image. More specifically, $Pr_i$ equals to the sum of values in the estimated density map.

\subsection{Datasets}

\textbf{ShanghaiTech dataset}. This is a large-scale crowd counting dataset, which contains 1198 images with 330,165 annotated heads. It is split into two parts: Part A has 482 images randomly collected from the Internet including 300 images for training and the rest for testing, and Part B contains 716 images taken from busy streets of metropolitan in Shanghai, and with 400 images for training and the remaining images for testing. We randomly crop 200 patches from each training image with the resolution of $ 192 \times 160 $. %During the test, we resize the size of the test image to be multiples of 8 as the size of the output of the network is 1/8 of that of the input. <- this is too detailed. No need to add this

\textbf{UCF\_CC\_50 dataset}. The UCF\_CC\_50 dataset has only 50 images captured from various perspectives, which is a very challenging counting dataset introduced by ~\cite{idrees2013multi}. On average, It contains 1280 persons per image ranging from 94 to 4543. We crop 60 patches from each image to train both methods as this dateset is too small, and followed by ~\cite{idrees2013multi}, 5-fold cross-validation is used to evaluate our proposed method.

\textbf{WorldExpo' 10}. The WorldExpo' 10 is the largest cross-scene crowd counting dataset introduced by~\cite{zhang2015cross, zhang2016data}. It consists of 1132 annotated video shot by 108 surveillance cameras from Shanghai 2010 WorldExpo. There are 3980 frames uniformly sampled from the videos sequences, where 3380 frames are used for training and the rest for testing. The number of pedestrians ranges from 1 to 220. Different from the above datasets, the region of interest (ROI) is provided for the images in the dataset. During data preprocessing, we mask each frame and its corresponding density map with ROI.

\begin{figure*}[h]
	\centering
    \includegraphics[scale=0.65]{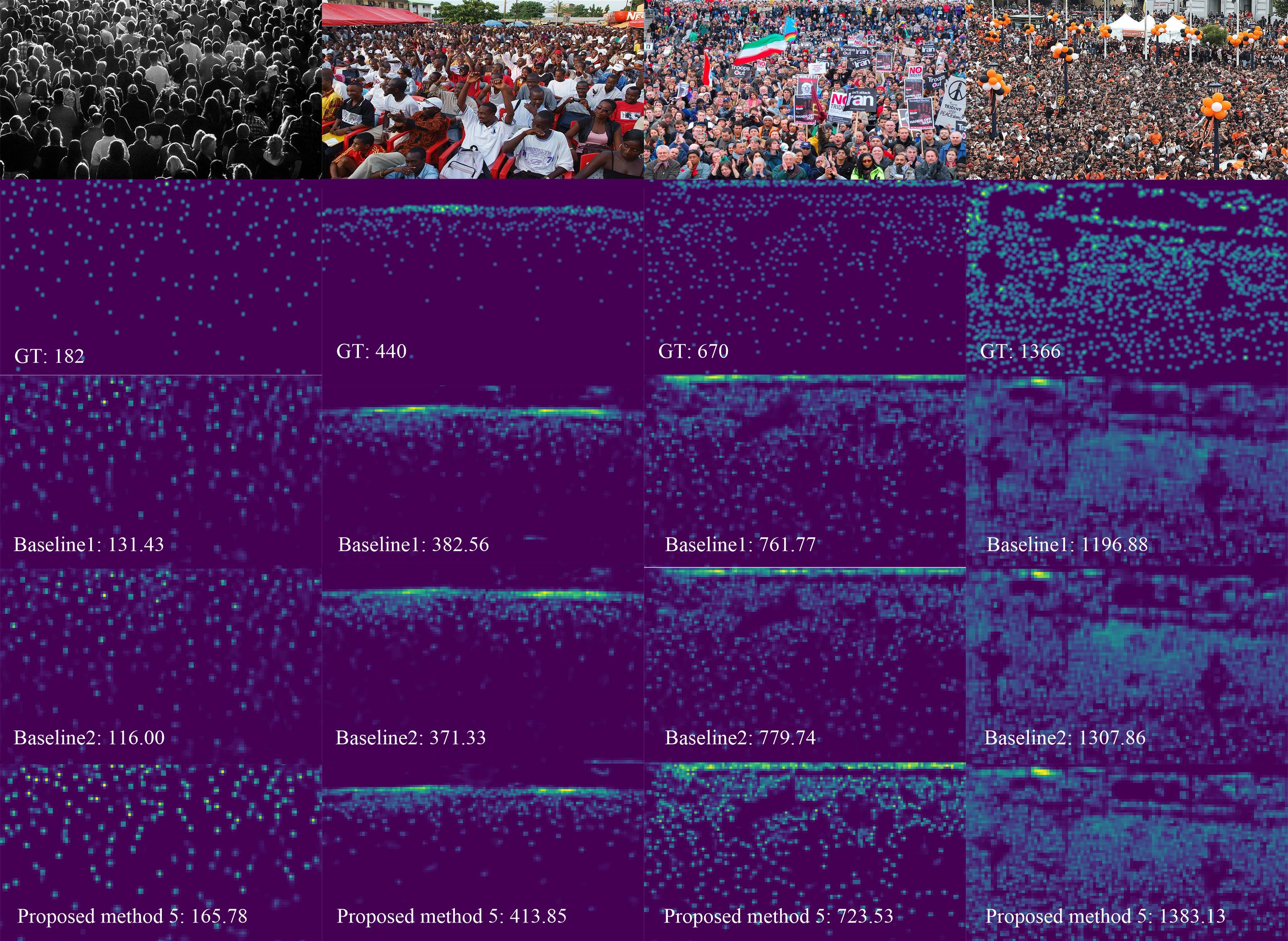}
	\caption{A comparison of the density map generated by our best-performed method and two baselines in ShanghaiTech Part A.}
	\label{fig:fig5}
\end{figure*} 

\subsection{Analysis of the proposed approaches} 
\begin{table}[t!]
% \caption{The algorithm performance of door dataset.}
\label{tab:door}
\caption{The experimental comparison on the baselines and five proposed methods on Shanghai Part A. Solution 1-5 corresponds to the architectures in Figure \ref{fig:fig3}. }
\begin{center}
\begin{tabular}{|c|c|c|}
\hline
 Method &MAE &MSE \\
 \hline
 baseline 1 &77.25 & 127.21\\
  \hline
 baseline 2 & 74.96 & 117.72  \\
 \hline
 Our proposed solution 1 &87.45&  128.81 \\
\hline
  Our proposed solution 2  & 69.77 &120.61 \\
\hline
  Our proposed solution 3 &76.66 &115.40 \\
 \hline
  Our proposed solution 4 &71.37 & 111.91 \\
 \hline
  Our proposed solution 5 &65.74 & 107.83  \\
\hline
\end{tabular}
\end{center}
% \caption{The comparison of our proposed methods on Shanghai Part A.}
\end{table}
%   Our proposed solution (3)&85.42  &130.62 \\

\begin{table}[t!]
% \caption{The algorithm performance of door dataset.}
\label{tab:door}
\caption{The experimental comparison on baselines, our proposed method and CSRNet on Shanghai Part A. }
\begin{center}
\begin{tabular}{|c|c|c|}
\hline
 Method &MAE &MSE \\
\hline
Baseline 3 & 73.66 & 120.26  \\
\hline
Our proposed solution 5 & 65.74 & 107.83  \\
\hline
CSRNet & 68.2 & 115.0  \\
\hline
Our model\_csr & 61.82 & 100.01  \\
  
\hline
\end{tabular}
\end{center}
% \caption{The comparison of our proposed methods on Shanghai Part A.}
\end{table}
%   Our proposed solution (3)&85.42  &130.62 \\

This paper proposes five different designs for the mask-aware density regressor. Its effectiveness will be examined in this subsection. We use solution 1-5 to denote the proposed architectures shown in Figure \ref{fig:fig1}. Besides these solutions, we also compare our method against three baselines (baseline 1-3) to verify the benefit of introducing mask-aware network design. The baselines are:
\begin{itemize}
\item \textbf{baseline 1} is a simply backbone subnetwork plus the density estimation branch as in solution 1-2. The purpose of presenting this baseline is to examine if adding mask branch and mask-aware density regressor can indeed lead to improvement. 
\item \textbf{baseline 2} is a deeper version of baseline 1. We notice that our solution 4 and 5 essentially use deeper networks for density regression. Thus it is fair to compare against a baseline with the comparable depth. 

\end{itemize}

The experiment results of the above methods are summarized in Table 1. From the results, we could make the following observations.

(1) The proposed solution 1 does not lead to the improved performance over baseline 1 which is comparable to it in terms of the network depth. On the contrary, it worsens the density estimation performance. In comparison, solution 2 leads to significant performance improvement. Comparing with baseline 1, it reduces the estimation error by 7 in MAE and MSE. This observation suggests that it is inappropriate to treat the mask prediction and density prediction independently. It is crucial to train those two tasks jointly.

(2) Somewhat surprising, the proposed solution 3 has no significant improvement. We postulate that it is due to the difficulty in optimizing the non-differentiable operator despite the fact that we have already approximated it by the straight-through estimator.

(3) Solution 4 also leads to an improved performance over baselines, although the improvement over its comparable method, baseline 2, is marginal. Note that solution 4 does not utilize the joint training strategy and the mask-aware density regressor will receive different mask inputs (ground truth and predicted) at the training and testing stage respectively. However, this limit does not prevent the method from gaining performance improvement. This may suggest that using convolutional layers to combine the mask prediction information is more robust than the elementwise product. 

(4) Our last solution 5 further achieves significant performance improvement over solution 4 and baseline 2. It reduces the MAE from 74.96 in baseline 2 to 65.74. This again shows the benefit of joint training and the power of using convolutional layers for information fusion. 

\subsection{Ablation study}

To have more insights into our proposed method, we conduct ablation studies of the proposed method on the part one of ShanghaiTech A dataset. The main studies and findings are presented below.

(1) To understand the effect of segmentation branch, we set a new baseline (baseline 3). This baseline uses identical network structure as our solution 5, but replaces the target of the mask prediction by density regression. In this way, the structure is similar to that in ~\cite{sam2018top}. This baseline is to verify whether the improvement of the proposed method merely comes from the architecture, or the mask prediction objective.

From Table II, it is not hard to conclude that our best-performed solution 5 still achieves significant improvement over baseline 3. Recall that the difference between solution 5 and baseline 3 is that the former adopts the mask prediction as the training objective. The performance discrepancy of these two methods suggests that using mask information could indeed benefit the density estimation. The improvement of our method does not solely come from the network structure. In Figure \ref{fig:fig5}, we also visualize the estimated density maps of our best-performed method and baseline approaches. From \ref{fig:fig5}, it is interesting to find that although the proposed approach gives more accurate count estimation, it does not provide a visibly better foreground/background separation than the baselines. This may suggest that the benefit of introducing the mask objective is not as simple as providing a better foreground/background separation. We postulate that the better performance achieved by our approach is due to that its density value estimation becomes more accurate with the guidance of the mask prediction.

(2) We also conduct a comparison experiment between binary cross entropy loss and focal loss. The result shows that the network with binary cross entropy loss can achieve almost the same performance: MAE: 66.08, MSE: 104.69 compared with that of the solution 5. So focal loss in this paper is a general setting for the mask branch.

(3) To compare the considered model with different backbones, we construct a new network with the same network structure in the solution 5 on top of a recent state-of-the-art network (CSRNet \cite{li2018csrnet}). To distinguish our proposed baseline, we term this network as our method\_csr for short.  As shown in Table II, we can see our method\_csr can achieve a promising improvement over the original CSRNet. To some extent, it indicates that a good baseline with the exploited network structure can boost the performance. Also note that the pre-trained model can be easily trained in a simple setting as shown in Sec. III(F) compared to our proposed model trained from  scratch.  We argue that the main benefits derives from the pre-trained VGG 16. Compared with CSRNet, our proposed baseline is more computationally efficient.

\begin{figure*}[h]
	\centering
    \includegraphics[scale=0.99]{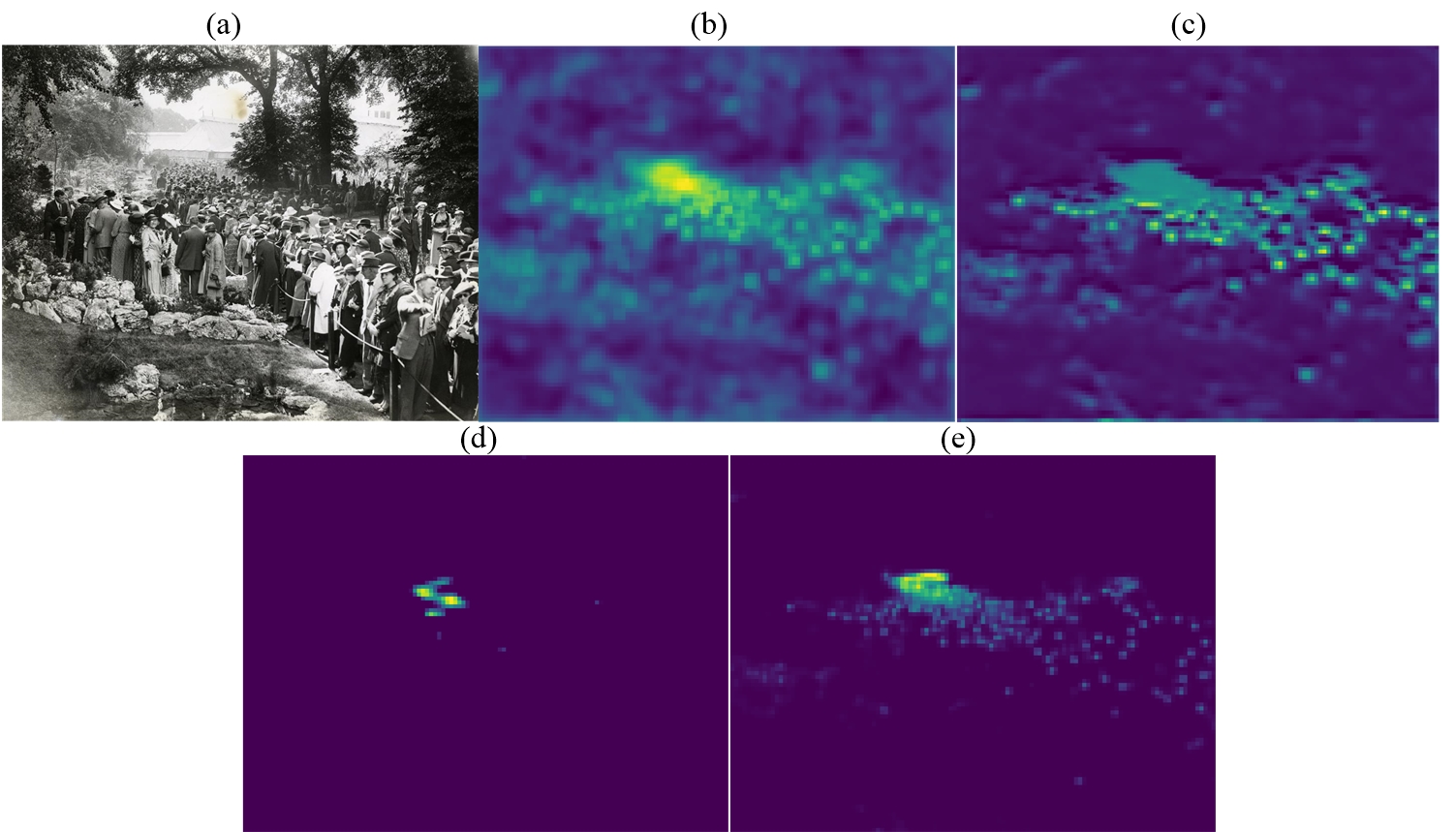}
	\caption{The visualization of feature maps in the mask branch. (a) is the input image; (b) is the output in segmentation branch; (c) is randomly selected feature map from the feedback convolution layers of the segmentation branch; (d) is randomly selected feature map after concatenating the feedback of segmentation branch; (e) is the final predicted density map.}
	\label{fig:fig6}
\end{figure*}

(4) To show the interactions among the ROI mask, input and density regression, we visualize the feature maps among those layers in Figure \ref{fig:fig6}. We use the test images(Figure 5(a)) in the part A of ShanghaiTech dataset.We find that there exist mask errors after the sigmoid layer of the segmentation branch as shown in Figure 5(b). We randomly selected one feature map (as shown in 'd1' Figure \ref{fig:fig3}) after feeding back the predicted mask posterior. Interestingly, from Figure 5(c), it can be seen that the errors in the mask prediction are suppressed in the sampled feature map. This suggests that the network has the capability of separating the error pattern at the mask prediction stage into different feature maps and potentially suppressing the error signal for density estimation. After the fusion of the two branches, each feature map in the regressor only focuses on a small part of the interest region shown in Figure 5(d). From the above discussion, we can see that even though there exist mask errors in the mask branch, they will not magnify in the next stage. Finally, we get a refined density map as shown in Figure 5(e). Here we argue that the mask error will not magnify in the next stage. 

% we can see that the mask errors are decreased when the information are feedback to the regression branch which shows a discriminable background and foreground. We argue that it mainly benefits from end-to-end training from the density regression.

\begin{figure}
	\centering
    \includegraphics[scale=0.5]{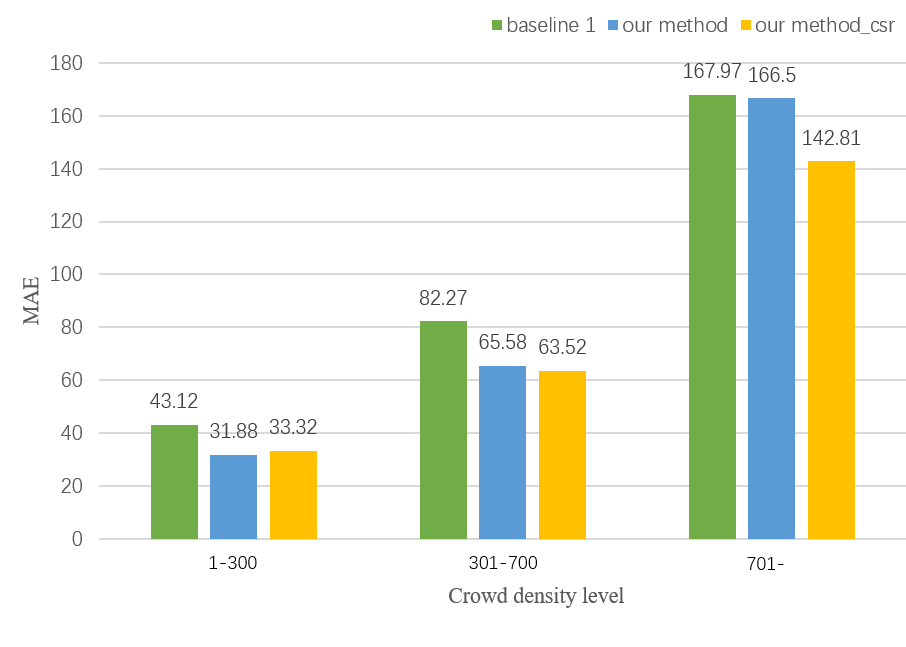}
	\caption{The average MAE of different density levels tested on ShanghaiTech Part A.}
	\label{fig:fig9}
\end{figure}

 (5) As is known, there are different density levels in the crowd. So we conduct the comparative experiment with three levels on ShanghaiTech Part A to show the improvement of our method. We split the density into three types of crowd: low crowd (1-300), middle crowd(301-700) and high crowd(700-). From Figure \ref{fig:fig9}, it is easily concluded that the proposed method achieves a promising result on the low and middle level of crowd. This is because the proposed segmentation branch has the ability to discriminate background and foreground. As for high-level crowd, it poses a challenging situation for most methods. The texture information of the crowd people are missing in those scenes so it is really hard to exact robust features for each head. As a result, we can not see clear promotion in this interval. As for the our method with pre-trained model, we can see that it has a similar improvement in low and middle crowds compared to the model with the proposed backbone while it also achieves a good result in high crowd. We conjecture that the pre-trained backbone has more prior knowledge to capture the texture information in high density level crowd than the model trained from scratch.  
 
\subsection{Comparison with the state-of-the-art}
We further compare our best-performed solution against the state-of-the-art results in various datasets. Firstly, we make a comparison on the Part A and Part B of the ShanghaiTech dataset. We compare our method against CC-Counting~\cite{zhang2015cross}, FCN~\cite{marsden2016fully},  MCNN,  TDF-net~\cite{sam2018top}, Switching-net~\cite{sam2017switching}, \cite{huang2018body} (BA-net for short), NetVLAD(multitask)~\cite{shi2018multiscale}, CP-CNN~\cite{sindagi2017generating} and CSRNet~\cite{li2018csrnet}. The results are summarized in Table 3. We can see that our method also achieves competitive results with the state-of-the-art methods~\cite{sindagi2017generating} and ~\cite{li2018csrnet} while keeping economic parameters. It is noted that the number of parameters of our proposed method is less than 5.1 million while the number in the CP-CNN is 68.4 million. So our method is more parameter economic and potentially more efficient. As for our method\_csr surpasses the two methods significantly in this dataset. Specifically, the MAE of Part A is 61.8, which outperforms that of the CSRNet by about 6.4. In terms of the MSE, our proposed method shows significant improvement over CSRNet by 13\%. In Part B, we also see that our method\_csr achieves 18.9\% in MAE and 16.9\% in MSE improvement compared to CSRNet on Part B. These results show the benefits of our proposed strategy in such a high variance scene.

In addition, we report the results of our approach on UCF\_CC\_50 dataset in Table 4. Our method obtains a 12.5 improvement in MAE over CP-CNN but is worse than CSRNet. We argue that the main reason lies in that the pre-trained model enjoys more prior information compared with our model trained from scratch especially in such a small dataset. Instead, our model\_csr armed with pre-trained VGG16 is superior to other models in MAE. It should be noted that it shows 20.7 and 38.2 improvement over CSRNet in terms of MAE and MSE, respectively. 

Finally, we present the results of our method on the WorldExpo' 10 dataset as shown in Table~\ref{tab.door2}. Our method with light weight achieves a relatively good performance which is on par with the state-of-the-art methods like TDF-net, NetVLAD(multitask), and classical methods MCNN and CC-counting but it is inferior to CSRNet and CP-CNN. Besides, our model\_csr precedes CSRNet and CP-CNN while obtaining the first place in S1, S2 and S5 scenes.

%Our method is slightly inferior to the CP-CNN and Switching-CNN, but our method trained from scratch and do not leverage the pre-trained model.

\begin{table}[t!]
\caption{The performance comparison on the ShanghaiTech dataset.}
\label{tab.door}
\begin{center}
\begin{tabular}{|c|cc|cc|}
\hline
       & \multicolumn{2}{|c|}{\textbf{Part A}}  & \multicolumn{2}{|c|}{\textbf{Part B}}\\
 Method & MAE & MSE &MAE & MSE\\
 
\hline
CC-Counting  & 181.8 & 277.7 & 32.0 & 49.8 \\
 \hline
FCN   & 126.5 & 173.5 & 23.8 & 33.1 \\
\hline
MCNN           & 110.2 &173.2  &26.4 &41.3 \\
\hline 
TDF-net   & 97.5 & 145.1 & 20.7 & 32.8 \\
\hline
Switching-net  & 90.4  & 135.0  &21.6 &33.4 \\
\hline
BA-CNN  & --& -- & 20.2 &35.6 \\
\hline
NetVLAD(multitask)   & 107.6 & 169.3 &21.4 & 33.9 \\
\hline
CP-CNN         &73.6  &  106.4 &20.1 &30.1 \\
\hline
CSRNet         &68.2  &  115.0 &10.6 &16.0 \\
\hline
Our method & 65.7 &107.8 &11.7 &16.4 \\
\hline
Our model\_csr & 61.8 & 100.0 & 8.6 & 13.3 \\

\hline
\end{tabular}
\end{center}
% \caption{The algorithm performance of ShanghaiTech dataset.}
\end{table}

\begin{table}[t!]
\caption{The performance comparison on the UCF\_CC\_50 dataset.}
\label{tab:door1}
\begin{center}
\begin{tabular}{|c|c|c|}
\hline
 Method      & MAE & MSE \\
\hline
CC-Counting    & 467.0 & 498.5 \\
\hline
FCN       & 338.6 & 424.5 \\
\hline
MCNN      & 377.6 &509.1 \\
\hline 
TDF-net   & 354.7 & 491.4  \\
\hline
Switching-net & 318.1  & 439.2  \\
\hline
BA-CNN  & 409.5 &563.7 \\
\hline
NetVLAD(multitask)   & 311.3 & 401.8 \\
\hline
CP-CNN         &295.8  &  320.9 \\
\hline
CSRNet         &266.1  &  397.5 \\
\hline
Our method &283.3 &411.6 \\
\hline
Our model\_csr &245.4 & 349.3\\
\hline
\end{tabular}
\end{center}
\end{table}

\begin{table}[t!]
\caption{The performance comparison on the WorldExpo'10 dataset.}
\label{tab.door2}
\begin{center}
\begin{tabular}{|c|c|c|c|c|c|c|}
\hline
 Method & S1 & S2 &S3 & S4 & S5 & Avg \\
 
\hline
CC-counting   & 9.8 & 14.1 & 14.3 & 22.2 & 3.7 & 12.82 \\
\hline
 MCNN           & 3.4 &20.6 &12.9 &13.0 & 8.1 & 11.6 \\
\hline 
TDF-net   & 2.7 & 23.4 & 10.7 & 17.6 & 3.3 & 11.54 \\
\hline
Switching-net  & 4.4  & 15.7  &10.0 &11.0 & 5.9 & 9.4 \\
\hline
BA-CNN  & 4.1& 21.7 & 11.9 &11.0 & 3.5 & 10.44 \\
\hline
NetVLAD(multitask)  &3.7 &  15.9 &10.2 &15.2 & 6.7 & 10.5 \\ 
\hline
CP-CNN         &2.9  &  14.7 &10.5 &10.4 & 5.8 & 8.86 \\
\hline
CSRNet        &2.9  &  11.5 &8.6 &16.6 &3.4 & 8.6 \\
\hline
 Our method &3.0 &16.7 &11.6 &12.5 &4.1 &9.58  \\
 \hline
Our model\_csr  &2.2 &11.5 &11.6 &13.9 &2.5 &8.34  \\
\hline
\end{tabular}
\end{center}
% \caption{The algorithm performance of ShanghaiTech dataset.}
\end{table}

\section{Conclusion}

In this paper, we address the crowd counting problem with deep neural networks. Our main discovery is the benefit of using a dedicated network branch to predict the foreground/background mask and incorporating mask prediction into density map estimation. We systematically study five different designs of the mask-aware density estimator and identify the best performed solution. Through the experimental validation, we show that the proposed scheme is effective and achieves the state-of-the-art crowd counting performance on various datasets.

\section*{Acknowledgment}

The authors would like to thank the editor and the anonymous reviewers for their valuable comments and constructive suggestions. This work is supported by the Scientific Research Foundation of Graduate School of Southeast University (No.YBJJ1768), the Postgraduate Research  \&  Practice  Innovation  Program  of  Jiangsu  Province (No.KYCX17\_0101),  the  National  Natural  Science  Foundation of China (No.61871123), Key Research and Development Program in Jiangsu Province (No.BE2016739) and a Project Funded by the Priority Academic Program Development of Jiangsu Higher Education Institutions.

\bibliographystyle{IEEEbib}
\bibliography{ieee_edit}

\begin{thebibliography}{10}

\bibitem{choi2014understanding}
Wongun Choi and Silvio Savarese,
\newblock ``Understanding collective activitiesof people from videos,''
\newblock {\em IEEE Transactions on Pattern Analysis and Machine Intelligence},
  vol. 36, no. 6, pp. 1242--1257, 2014.

\bibitem{shao2017crowded}
Jing Shao, Chen~Change Loy, Kai Kang, and Xiaogang Wang,
\newblock ``Crowded scene understanding by deeply learned volumetric slices,''
\newblock {\em IEEE Transactions on Circuits and Systems for Video Technology},
  vol. 27, no. 3, pp. 613--623, 2017.

\bibitem{huang2018body}
Siyu Huang, Xi~Li, Zhongfei Zhang, Fei Wu, Shenghua Gao, Rongrong Ji, and
  Junwei Han,
\newblock ``Body structure aware deep crowd counting,''
\newblock {\em IEEE Transactions on Image Processing}, vol. 27, no. 3, pp.
  1049--1059, 2018.

\bibitem{li2018deep}
Yuke Li,
\newblock ``A deep spatiotemporal perspective for understanding crowd
  behavior,''
\newblock {\em IEEE Transactions on Multimedia}, 2018.

\bibitem{zhang2016data}
Cong Zhang, Kai Kang, Hongsheng Li, Xiaogang Wang, Rong Xie, and Xiaokang Yang,
\newblock ``Data-driven crowd understanding: a baseline for a large-scale crowd
  dataset,''
\newblock {\em IEEE Transactions on Multimedia}, vol. 18, no. 6, pp.
  1048--1061, 2016.

\bibitem{sindagi2018survey}
Vishwanath~A Sindagi and Vishal~M Patel,
\newblock ``A survey of recent advances in cnn-based single image crowd
  counting and density estimation,''
\newblock {\em Pattern Recognition Letters}, vol. 107, pp. 3--16, 2018.

\bibitem{ren2015faster}
Shaoqing Ren, Kaiming He, Ross Girshick, and Jian Sun,
\newblock ``Faster r-cnn: Towards real-time object detection with region
  proposal networks,''
\newblock in {\em Advances in Neural Information Processing Systems}, 2015, pp.
  91--99.

\bibitem{liu2016ssd}
Wei Liu, Dragomir Anguelov, Dumitru Erhan, Christian Szegedy, Scott Reed,
  Cheng-Yang Fu, and Alexander~C Berg,
\newblock ``Ssd: Single shot multibox detector,''
\newblock in {\em European Conference on Computer Vision}. Springer, 2016, pp.
  21--37.

\bibitem{redmon2016you}
Joseph Redmon, Santosh Divvala, Ross Girshick, and Ali Farhadi,
\newblock ``You only look once: Unified, real-time object detection,''
\newblock in {\em IEEE Conference on Computer Vision and Pattern Recognition},
  2016, pp. 779--788.

\bibitem{li2018multistage}
Jianan Li, Xiaodan Liang, Jianshu Li, Yunchao Wei, Tingfa Xu, Jiashi Feng, and
  Shuicheng Yan,
\newblock ``Multistage object detection with group recursive learning,''
\newblock {\em IEEE Transactions on Multimedia}, vol. 20, no. 7, pp.
  1645--1655, 2018.

\bibitem{zhang2016single}
Yingying Zhang, Desen Zhou, Siqin Chen, Shenghua Gao, and Yi~Ma,
\newblock ``Single-image crowd counting via multi-column convolutional neural
  network,''
\newblock in {\em IEEE Conference on Computer Vision and Pattern Recognition},
  2016, pp. 589--597.

\bibitem{wang2015deep}
Chuan Wang, Hua Zhang, Liang Yang, Si~Liu, and Xiaochun Cao,
\newblock ``Deep people counting in extremely dense crowds,''
\newblock in {\em ACM International Conference on Multimedia}. ACM, 2015, pp.
  1299--1302.

\bibitem{sam2017switching}
Deepak~Babu Sam, Shiv Surya, and R~Venkatesh Babu,
\newblock ``Switching convolutional neural network for crowd counting,''
\newblock in {\em IEEE Conference on Computer Vision and Pattern Recognition},
  2017, vol.~1, p.~6.

\bibitem{ge2009marked}
Weina Ge and Robert~T Collins,
\newblock ``Marked point processes for crowd counting,''
\newblock in {\em IEEE Conference on Computer Vision and Pattern Recognition}.
  IEEE, 2009, pp. 2913--2920.

\bibitem{ge2012vision}
Weina Ge, Robert~T Collins, and R~Barry Ruback,
\newblock ``Vision-based analysis of small groups in pedestrian crowds,''
\newblock {\em IEEE Transactions on Pattern Analysis and Machine Intelligence},
  vol. 34, no. 5, pp. 1003--1016, 2012.

\bibitem{idrees2015detecting}
Haroon Idrees, Khurram Soomro, and Mubarak Shah,
\newblock ``Detecting humans in dense crowds using locally-consistent scale
  prior and global occlusion reasoning,''
\newblock {\em IEEE Transactions on Pattern Analysis and Machine Intelligence},
  vol. 37, no. 10, pp. 1986--1998, 2015.

\bibitem{onoro2016towards}
Daniel Onoro-Rubio and Roberto~J L{\'o}pez-Sastre,
\newblock ``Towards perspective-free object counting with deep learning,''
\newblock in {\em European Conference on Computer Vision}. Springer, 2016, pp.
  615--629.

\bibitem{loy2013crowd}
Chen~Change Loy, Ke~Chen, Shaogang Gong, and Tao Xiang,
\newblock ``Crowd counting and profiling: Methodology and evaluation,''
\newblock in {\em Modeling, Simulation and Visual Analysis of Crowds}, pp.
  347--382. Springer, 2013.

\bibitem{tripathi2018convolutional}
Gaurav Tripathi, Kuldeep Singh, and Dinesh~Kumar Vishwakarma,
\newblock ``Convolutional neural networks for crowd behaviour analysis: a
  survey,''
\newblock {\em The Visual Computer}, pp. 1--24, 2018.

\bibitem{viola2004robust}
Paul Viola and Michael~J Jones,
\newblock ``Robust real-time face detection,''
\newblock {\em International Journal of Computer Vision}, vol. 57, no. 2, pp.
  137--154, 2004.

\bibitem{dalal2005histograms}
Navneet Dalal and Bill Triggs,
\newblock ``Histograms of oriented gradients for human detection,''
\newblock in {\em IEEE Conference on Computer Vision and Pattern Recognition}.
  IEEE, 2005, vol.~1, pp. 886--893.

\bibitem{enzweiler2008monocular}
Markus Enzweiler and Dariu~M Gavrila,
\newblock ``Monocular pedestrian detection: Survey and experiments,''
\newblock {\em IEEE Transactions on Pattern Analysis and Machine Intelligence},
  , no. 12, pp. 2179--2195, 2008.

\bibitem{zhao2008segmentation}
Tao Zhao, Ram Nevatia, and Bo~Wu,
\newblock ``Segmentation and tracking of multiple humans in crowded
  environments,''
\newblock {\em IEEE Transactions on Pattern Analysis and Machine Intelligence},
  vol. 30, no. 7, pp. 1198--1211, 2008.

\bibitem{yang2003counting}
Danny~B Yang, Leonidas~J Guibas, et~al.,
\newblock ``Counting people in crowds with a real-time network of simple image
  sensors,''
\newblock in {\em null}. IEEE, 2003, p. 122.

\bibitem{ren2017faster}
Shaoqing Ren, Kaiming He, Ross Girshick, and Jian Sun,
\newblock ``Faster r-cnn: towards real-time object detection with region
  proposal networks,''
\newblock {\em IEEE Transactions on Pattern Analysis and Machine Intelligence},
  , no. 6, pp. 1137--1149, 2017.

\bibitem{lempitsky2010learning}
Victor Lempitsky and Andrew Zisserman,
\newblock ``Learning to count objects in images,''
\newblock in {\em Advances in Neural Information Processing Systems}, 2010, pp.
  1324--1332.

\bibitem{chan2012counting}
Antoni~B Chan and Nuno Vasconcelos,
\newblock ``Counting people with low-level features and bayesian regression,''
\newblock {\em IEEE Transactions on Image Processing}, vol. 21, no. 4, pp.
  2160--2177, 2012.

\bibitem{ma2013crossing}
Zheng Ma and Antoni~B Chan,
\newblock ``Crossing the line: Crowd counting by integer programming with local
  features,''
\newblock in {\em IEEE Conference on Computer Vision and Pattern Recognition},
  2013, pp. 2539--2546.

\bibitem{idrees2013multi}
Haroon Idrees, Imran Saleemi, Cody Seibert, and Mubarak Shah,
\newblock ``Multi-source multi-scale counting in extremely dense crowd
  images,''
\newblock in {\em IEEE Conference on Computer Vision and Pattern Recognition},
  2013, pp. 2547--2554.

\bibitem{chen2013cumulative}
Ke~Chen, Shaogang Gong, Tao Xiang, and Chen Change~Loy,
\newblock ``Cumulative attribute space for age and crowd density estimation,''
\newblock in {\em IEEE Conference on Computer Vision and Pattern Recognition},
  2013, pp. 2467--2474.

\bibitem{zhang2015cross}
Cong Zhang, Hongsheng Li, Xiaogang Wang, and Xiaokang Yang,
\newblock ``Cross-scene crowd counting via deep convolutional neural
  networks,''
\newblock in {\em IEEE Conference on Computer Vision and Pattern Recognition},
  2015, pp. 833--841.

\bibitem{sindagi2017generating}
Vishwanath~A Sindagi and Vishal~M Patel,
\newblock ``Generating high-quality crowd density maps using contextual pyramid
  cnns,''
\newblock in {\em IEEE International Conference on Computer Vision}. IEEE,
  2017, pp. 1879--1888.

\bibitem{ranjan2018iterative}
Viresh Ranjan, Hieu Le, and Minh Hoai,
\newblock ``Iterative crowd counting,''
\newblock {\em arXiv preprint arXiv:1807.09959}, 2018.

\bibitem{shen2018crowd}
Zan Shen, Yi~Xu, Bingbing Ni, Minsi Wang, Jianguo Hu, and Xiaokang Yang,
\newblock ``Crowd counting via adversarial cross-scale consistency pursuit,''
\newblock in {\em IEEE Conference on Computer Vision and Pattern Recognition},
  2018, pp. 5245--5254.

\bibitem{shi2018crowd}
Zenglin Shi, Le~Zhang, Yun Liu, Xiaofeng Cao, Yangdong Ye, Ming-Ming Cheng, and
  Guoyan Zheng,
\newblock ``Crowd counting with deep negative correlation learning,''
\newblock in {\em IEEE Conference on Computer Vision and Pattern Recognition},
  2018, pp. 5382--5390.

\bibitem{li2018csrnet}
Yuhong Li, Xiaofan Zhang, and Deming Chen,
\newblock ``Csrnet: Dilated convolutional neural networks for understanding the
  highly congested scenes,''
\newblock in {\em IEEE Conference on Computer Vision and Pattern Recognition},
  2018, pp. 1091--1100.

\bibitem{sam2018top}
Deepak~Babu Sam and R~Venkatesh Babu,
\newblock ``Top-down feedback for crowd counting convolutional neural
  network,''
\newblock {\em arXiv preprint arXiv:1807.08881}, 2018.

\bibitem{he2017mask}
Kaiming He, Georgia Gkioxari, Piotr Doll{\'a}r, and Ross Girshick,
\newblock ``Mask r-cnn,''
\newblock in {\em IEEE International Conference on Computer Vision}. IEEE,
  2017, pp. 2980--2988.

\bibitem{lyu2018mask}
Pengyuan Lyu, Minghui Liao, Cong Yao, Wenhao Wu, and Xiang Bai,
\newblock ``Mask textspotter: An end-to-end trainable neural network for
  spotting text with arbitrary shapes,''
\newblock {\em arXiv preprint arXiv:1807.02242}, 2018.

\bibitem{szegedy2016rethinking}
Christian Szegedy, Vincent Vanhoucke, Sergey Ioffe, Jon Shlens, and Zbigniew
  Wojna,
\newblock ``Rethinking the inception architecture for computer vision,''
\newblock in {\em IEEE Conference on Computer Vision and Pattern Recognition},
  2016, pp. 2818--2826.

\bibitem{lin2018focal}
Tsung-Yi Lin, Priyal Goyal, Ross Girshick, Kaiming He, and Piotr Doll{\'a}r,
\newblock ``Focal loss for dense object detection,''
\newblock {\em IEEE Transactions on Pattern Analysis and Machine Intelligenc},
  2018.

\bibitem{bengio2013estimating}
Yoshua Bengio, Nicholas L{\'e}onard, and Aaron Courville,
\newblock ``Estimating or propagating gradients through stochastic neurons for
  conditional computation,''
\newblock {\em arXiv preprint arXiv:1308.3432}, 2013.

\bibitem{marsden2016fully}
Mark Marsden, Kevin McGuinness, Suzanne Little, and Noel~E O'Connor,
\newblock ``Fully convolutional crowd counting on highly congested scenes,''
\newblock {\em arXiv preprint arXiv:1612.00220}, 2016.

\bibitem{shi2018multiscale}
Zenglin Shi, Le~Zhang, Yibo Sun, and Yangdong Ye,
\newblock ``Multiscale multitask deep netvlad for crowd counting,''
\newblock {\em IEEE Transactions on Industrial Informatics}, vol. 14, no. 11,
  pp. 4953--4962, 2018.

\end{thebibliography}


@article{shi2018multiscale,
  title={Multiscale multitask deep NetVLAD for crowd counting},
  author={Shi, Zenglin and Zhang, Le and Sun, Yibo and Ye, Yangdong},
  journal={IEEE Transactions on Industrial Informatics},
  volume={14},
  number={11},
  pages={4953--4962},
  year={2018},
  publisher={IEEE}
}




@incollection{loy2013crowd,
  title={Crowd counting and profiling: Methodology and evaluation},
  author={Loy, Chen Change and Chen, Ke and Gong, Shaogang and Xiang, Tao},
  booktitle={Modeling, Simulation and Visual Analysis of Crowds},
  pages={347--382},
  year={2013},
  publisher={Springer}
}

@article{sam2018top,
  title={Top-down feedback for crowd counting convolutional neural network},
  author={Sam, Deepak Babu and Babu, R Venkatesh},
  journal={arXiv preprint arXiv:1807.08881},
  year={2018}
}

@article{huang2018body,
  title={Body structure aware deep crowd counting},
  author={Huang, Siyu and Li, Xi and Zhang, Zhongfei and Wu, Fei and Gao, Shenghua and Ji, Rongrong and Han, Junwei},
  journal={IEEE Transactions on Image Processing},
  volume={27},
  number={3},
  pages={1049--1059},
  year={2018},
  publisher={IEEE}
}

@article{zhao2008segmentation,
  title={Segmentation and tracking of multiple humans in crowded environments},
  author={Zhao, Tao and Nevatia, Ram and Wu, Bo},
  journal={IEEE Transactions on Pattern Analysis and Machine Intelligence},
  volume={30},
  number={7},
  pages={1198--1211},
  year={2008},
  publisher={IEEE}
}

@inproceedings{shi2018crowd,
  title={Crowd Counting With Deep Negative Correlation Learning},
  author={Shi, Zenglin and Zhang, Le and Liu, Yun and Cao, Xiaofeng and Ye, Yangdong and Cheng, Ming-Ming and Zheng, Guoyan},
  booktitle={IEEE Conference on Computer Vision and Pattern Recognition},
  pages={5382--5390},
  year={2018}
}

@inproceedings{yang2003counting,
  title={Counting people in crowds with a real-time network of simple image sensors},
  author={Yang, Danny B and Guibas, Leonidas J and others},
  booktitle={null},
  pages={122},
  year={2003},
  organization={IEEE}
}

@inproceedings{dalal2005histograms,
  title={Histograms of oriented gradients for human detection},
  author={Dalal, Navneet and Triggs, Bill},
  booktitle={IEEE Conference on Computer Vision and Pattern Recognition},
  volume={1},
  pages={886--893},
  year={2005},
  organization={IEEE}
}

@article{viola2004robust,
  title={Robust real-time face detection},
  author={Viola, Paul and Jones, Michael J},
  journal={International Journal of Computer Vision},
  volume={57},
  number={2},
  pages={137--154},
  year={2004},
  publisher={Springer}
}

@inproceedings{ge2009marked,
  title={Marked point processes for crowd counting},
  author={Ge, Weina and Collins, Robert T},
  booktitle={IEEE Conference on Computer Vision and Pattern Recognition},
  pages={2913--2920},
  year={2009},
  organization={IEEE}
}

@article{ge2012vision,
  title={Vision-based analysis of small groups in pedestrian crowds},
  author={Ge, Weina and Collins, Robert T and Ruback, R Barry},
  journal={IEEE Transactions on Pattern Analysis and Machine Intelligence},
  volume={34},
  number={5},
  pages={1003--1016},
  year={2012},
  publisher={IEEE}
}

@article{idrees2015detecting,
  title={Detecting humans in dense crowds using locally-consistent scale prior and global occlusion reasoning},
  author={Idrees, Haroon and Soomro, Khurram and Shah, Mubarak},
  journal={IEEE Transactions on Pattern Analysis and Machine Intelligence},
  volume={37},
  number={10},
  pages={1986--1998},
  year={2015},
  publisher={IEEE}
}

@article{shao2017crowded,
  title={Crowded scene understanding by deeply learned volumetric slices},
  author={Shao, Jing and Loy, Chen Change and Kang, Kai and Wang, Xiaogang},
  journal={IEEE Transactions on Circuits and Systems for Video Technology},
  volume={27},
  number={3},
  pages={613--623},
  year={2017},
  publisher={IEEE}
}

@inproceedings{onoro2016towards,
  title={Towards perspective-free object counting with deep learning},
  author={Onoro-Rubio, Daniel and L{\'o}pez-Sastre, Roberto J},
  booktitle={European Conference on Computer Vision},
  pages={615--629},
  year={2016},
  organization={Springer}
}

@inproceedings{shen2018crowd,
  title={Crowd Counting via Adversarial Cross-Scale Consistency Pursuit},
  author={Shen, Zan and Xu, Yi and Ni, Bingbing and Wang, Minsi and Hu, Jianguo and Yang, Xiaokang},
  booktitle={IEEE Conference on Computer Vision and Pattern Recognition},
  pages={5245--5254},
  year={2018}
}

@article{wu2018collective,
  title={Collective Density Clustering for Coherent Motion Detection},
  author={Wu, Yunpeng and Ye, Yangdong and Zhao, Chenyang and Shi, Zenglin},
  journal={IEEE Transactions on Multimedia},
  volume={20},
  number={6},
  pages={1418--1431},
  year={2018},
  publisher={IEEE}
}

@article{li2018deep,
  title={A Deep Spatiotemporal Perspective for Understanding Crowd Behavior},
  author={Li, Yuke},
  journal={IEEE Transactions on Multimedia},
  year={2018},
  publisher={IEEE}
}

@article{li2018multistage,
  title={Multistage Object Detection With Group Recursive Learning},
  author={Li, Jianan and Liang, Xiaodan and Li, Jianshu and Wei, Yunchao and Xu, Tingfa and Feng, Jiashi and Yan, Shuicheng},
  journal={IEEE Transactions on Multimedia},
  volume={20},
  number={7},
  pages={1645--1655},
  year={2018},
  publisher={IEEE}
}


@article{choi2014understanding,
  title={Understanding collective activitiesof people from videos},
  author={Choi, Wongun and Savarese, Silvio},
  journal={IEEE Transactions on Pattern Analysis and Machine Intelligence},
  volume={36},
  number={6},
  pages={1242--1257},
  year={2014},
  publisher={IEEE}
}


@inproceedings{li2018csrnet,
  title={CSRNet: Dilated convolutional neural networks for understanding the highly congested scenes},
  author={Li, Yuhong and Zhang, Xiaofan and Chen, Deming},
  booktitle={IEEE Conference on Computer Vision and Pattern Recognition},
  pages={1091--1100},
  year={2018}
}


@inproceedings{chen2013cumulative,
  title={Cumulative attribute space for age and crowd density estimation},
  author={Chen, Ke and Gong, Shaogang and Xiang, Tao and Change Loy, Chen},
  booktitle={IEEE Conference on Computer Vision and Pattern Recognition},
  pages={2467--2474},
  year={2013}
}


@article{zhang2016data,
  title={Data-driven crowd understanding: a baseline for a large-scale crowd dataset},
  author={Zhang, Cong and Kang, Kai and Li, Hongsheng and Wang, Xiaogang and Xie, Rong and Yang, Xiaokang},
  journal={IEEE Transactions on Multimedia},
  volume={18},
  number={6},
  pages={1048--1061},
  year={2016},
  publisher={IEEE}
}


@inproceedings{idrees2013multi,
  title={Multi-source multi-scale counting in extremely dense crowd images},
  author={Idrees, Haroon and Saleemi, Imran and Seibert, Cody and Shah, Mubarak},
  booktitle={IEEE Conference on Computer Vision and Pattern Recognition},
  pages={2547--2554},
  year={2013}
}

@article{ren2017faster,
  title={Faster R-CNN: towards real-time object detection with region proposal networks},
  author={Ren, Shaoqing and He, Kaiming and Girshick, Ross and Sun, Jian},
  journal={IEEE Transactions on Pattern Analysis and Machine Intelligence},
  number={6},
  pages={1137--1149},
  year={2017},
  publisher={IEEE}
}

@inproceedings{redmon2016you,
  title={You only look once: Unified, real-time object detection},
  author={Redmon, Joseph and Divvala, Santosh and Girshick, Ross and Farhadi, Ali},
  booktitle={IEEE Conference on Computer Vision and Pattern Recognition},
  pages={779--788},
  year={2016}
}

@inproceedings{liu2016ssd,
  title={Ssd: Single shot multibox detector},
  author={Liu, Wei and Anguelov, Dragomir and Erhan, Dumitru and Szegedy, Christian and Reed, Scott and Fu, Cheng-Yang and Berg, Alexander C},
  booktitle={European Conference on Computer Vision},
  pages={21--37},
  year={2016},
  organization={Springer}
}


@article{tripathi2018convolutional,
  title={Convolutional neural networks for crowd behaviour analysis: a survey},
  author={Tripathi, Gaurav and Singh, Kuldeep and Vishwakarma, Dinesh Kumar},
  journal={The Visual Computer},
  pages={1--24},
  year={2018},
  publisher={Springer}
}


@article{enzweiler2008monocular,
  title={Monocular pedestrian detection: Survey and experiments},
  author={Enzweiler, Markus and Gavrila, Dariu M},
  journal={IEEE Transactions on Pattern Analysis and Machine Intelligence},
  number={12},
  pages={2179--2195},
  year={2008},
  publisher={IEEE}
}

@article{sindagi2018survey,
  title={A survey of recent advances in cnn-based single image crowd counting and density estimation},
  author={Sindagi, Vishwanath A and Patel, Vishal M},
  journal={Pattern Recognition Letters},
  volume={107},
  pages={3--16},
  year={2018},
  publisher={Elsevier}
}

@inproceedings{lempitsky2010learning,
  title={Learning to count objects in images},
  author={Lempitsky, Victor and Zisserman, Andrew},
  booktitle={Advances in Neural Information Processing Systems},
  pages={1324--1332},
  year={2010}
}


@article{bengio2013estimating,
  title={Estimating or propagating gradients through stochastic neurons for conditional computation},
  author={Bengio, Yoshua and L{\'e}onard, Nicholas and Courville, Aaron},
  journal={arXiv preprint arXiv:1308.3432},
  year={2013}
}

@article{chan2012counting,
  title={Counting people with low-level features and Bayesian regression},
  author={Chan, Antoni B and Vasconcelos, Nuno},
  journal={IEEE Transactions on Image Processing},
  volume={21},
  number={4},
  pages={2160--2177},
  year={2012},
  publisher={IEEE}
}

@inproceedings{ma2013crossing,
  title={Crossing the line: Crowd counting by integer programming with local features},
  author={Ma, Zheng and Chan, Antoni B},
  booktitle={IEEE Conference on Computer Vision and Pattern Recognition},
  pages={2539--2546},
  year={2013}
}

@inproceedings{shang2016end,
  title={End-to-end crowd counting via joint learning local and global count},
  author={Shang, Chong and Ai, Haizhou and Bai, Bo},
  booktitle={IEEE International Conference on Image Processing},
  pages={1215--1219},
  year={2016},
  organization={IEEE}
}

@inproceedings{zhang2015cross,
  title={Cross-scene crowd counting via deep convolutional neural networks},
  author={Zhang, Cong and Li, Hongsheng and Wang, Xiaogang and Yang, Xiaokang},
  booktitle={IEEE Conference on Computer Vision and Pattern Recognition},
  pages={833--841},
  year={2015}
}

@inproceedings{wang2015deep,
  title={Deep people counting in extremely dense crowds},
  author={Wang, Chuan and Zhang, Hua and Yang, Liang and Liu, Si and Cao, Xiaochun},
  booktitle={ACM International Conference on Multimedia},
  pages={1299--1302},
  year={2015},
  organization={ACM}
}


@inproceedings{zhang2016single,
  title={Single-image crowd counting via multi-column convolutional neural network},
  author={Zhang, Yingying and Zhou, Desen and Chen, Siqin and Gao, Shenghua and Ma, Yi},
  booktitle={IEEE Conference on Computer Vision and Pattern Recognition},
  pages={589--597},
  year={2016}
}

@inproceedings{sam2017switching,
  title={Switching convolutional neural network for crowd counting},
  author={Sam, Deepak Babu and Surya, Shiv and Babu, R Venkatesh},
  booktitle={IEEE Conference on Computer Vision and Pattern Recognition},
  volume={1},
  number={3},
  pages={6},
  year={2017}
}

@inproceedings{sindagi2017generating,
  title={Generating high-quality crowd density maps using contextual pyramid cnns},
  author={Sindagi, Vishwanath A and Patel, Vishal M},
  booktitle={IEEE International Conference on Computer Vision},
  pages={1879--1888},
  year={2017},
  organization={IEEE}
}

@inproceedings{he2017mask,
  title={Mask r-cnn},
  author={He, Kaiming and Gkioxari, Georgia and Doll{\'a}r, Piotr and Girshick, Ross},
  booktitle={IEEE International Conference on Computer Vision},
  pages={2980--2988},
  year={2017},
  organization={IEEE}
}


@article{lyu2018mask,
  title={Mask TextSpotter: An End-to-End Trainable Neural Network for Spotting Text with Arbitrary Shapes},
  author={Lyu, Pengyuan and Liao, Minghui and Yao, Cong and Wu, Wenhao and Bai, Xiang},
  journal={arXiv preprint arXiv:1807.02242},
  year={2018}
}

@inproceedings{ren2015faster,
  title={Faster r-cnn: Towards real-time object detection with region proposal networks},
  author={Ren, Shaoqing and He, Kaiming and Girshick, Ross and Sun, Jian},
  booktitle={Advances in Neural Information Processing Systems},
  pages={91--99},
  year={2015}
}


@inproceedings{chen2012feature,
  title={Feature mining for localised crowd counting.},
  author={Chen, Ke and Loy, Chen Change and Gong, Shaogang and Xiang, Tony},
  booktitle={British Machine Vision Conference},
  volume={1},
  number={2},
  pages={3},
  year={2012}
}

@article{marsden2016fully,
  title={Fully convolutional crowd counting on highly congested scenes},
  author={Marsden, Mark and McGuinness, Kevin and Little, Suzanne and O'Connor, Noel E},
  journal={arXiv preprint arXiv:1612.00220},
  year={2016}
}

@article{huang2018stacked,
  title={Stacked Pooling: Improving Crowd Counting by Boosting Scale Invariance},
  author={Huang, Siyu and Li, Xi and Cheng, Zhi-Qi and Zhang, Zhongfei and Hauptmann, Alexander},
  journal={arXiv preprint arXiv:1808.07456},
  year={2018}
}

@inproceedings{pham2015count,
  title={Count forest: Co-voting uncertain number of targets using random forest for crowd density estimation},
  author={Pham, Viet-Quoc and Kozakaya, Tatsuo and Yamaguchi, Osamu and Okada, Ryuzo},
  booktitle={IEEE International Conference on Computer Vision},
  pages={3253--3261},
  year={2015}
}
@inproceedings{wang2016fast,
  title={Fast visual object counting via example-based density estimation},
  author={Wang, Yi and Zou, Yuexian},
  booktitle={IEEE International Conference on Image Processing},
  pages={3653--3657},
  year={2016},
  organization={IEEE}
}

@article{ranjan2018iterative,
  title={Iterative crowd counting},
  author={Ranjan, Viresh and Le, Hieu and Hoai, Minh},
  journal={arXiv preprint arXiv:1807.09959},
  year={2018}
}

@article{kumagai2017mixture,
  title={Mixture of Counting CNNs: Adaptive Integration of CNNs Specialized to Specific Appearance for Crowd Counting},
  author={Kumagai, Shohei and Hotta, Kazuhiro and Kurita, Takio},
  journal={arXiv preprint arXiv:1703.09393},
  year={2017}
}

@article{sheng2016crowd,
  title={Crowd counting via weighted vlad on dense attribute feature maps},
  author={Sheng, Biyun and Shen, Chunhua and Lin, Guosheng and Li, Jun and Yang, Wankou and Sun, Changyin},
  journal={IEEE Transactions on Circuits and Systems for Video Technology},
  year={2016},
  publisher={IEEE}
}

@inproceedings{szegedy2016rethinking,
  title={Rethinking the inception architecture for computer vision},
  author={Szegedy, Christian and Vanhoucke, Vincent and Ioffe, Sergey and Shlens, Jon and Wojna, Zbigniew},
  booktitle={IEEE Conference on Computer Vision and Pattern Recognition},
  pages={2818--2826},
  year={2016}
}
@article{lin2018focal,
  title={Focal loss for dense object detection},
  author={Lin, Tsung-Yi and Goyal, Priyal and Girshick, Ross and He, Kaiming and Doll{\'a}r, Piotr},
  journal={IEEE Transactions on Pattern Analysis and Machine Intelligenc},
  year={2018},
  publisher={IEEE}
}

% \end{IEEEbiography}

\end{document}